\documentclass[letterpaper]{article}
\usepackage{emnlp2016}
\usepackage{times}
\usepackage{url}
\usepackage{latexsym}
\usepackage{amssymb}
\usepackage{graphicx}
\usepackage{float}
\usepackage{booktabs}
\usepackage{graphics}
\usepackage[fleqn]{amsmath}
\usepackage{multirow}
\emnlpfinalcopy
\usepackage{color}

\title{Are Word Embedding-based Features Useful for Sarcasm Detection?}
\author{\begin{tabular}{ccc}
Aditya Joshi$^{1,2,3}$& Vaibhav Tripathi$^{1}$ &Kevin Patel$^{1}$
\end{tabular}\\
\begin{tabular}{cc}
\textbf{Pushpak Bhattacharyya$^{1}$} & \textbf{Mark Carman$^{2}$}\\
\end{tabular}\\
\begin{tabular}{ccc}
\multicolumn{3}{c}{$^{1}$Indian Institute of Technology Bombay, India}\\
\multicolumn{3}{c}{$^{2}$Monash University, Australia}\\
\multicolumn{3}{c}{$^{3}$IITB-Monash Research Academy, India}\\
\multicolumn{3}{c}{\tt \{adityaj,kevin.patel,pb\}@cse.iitb.ac.in, mark.carman@monash.edu}
\end{tabular}
}
\begin{document}
\maketitle
\begin{abstract}
This paper makes a simple increment to state-of-the-art in sarcasm detection research. Existing approaches are unable to capture subtle forms of context incongruity which lies at the heart of sarcasm. We explore if prior work can be enhanced using semantic similarity/discordance between word embeddings.  We augment word embedding-based features to four feature sets reported in the past. We also experiment with four types of word embeddings. We observe an improvement in sarcasm detection, irrespective of the word embedding used or the original feature set to which our features are augmented. For example, this augmentation results in an improvement in F-score of around 4\% for three out of these four feature sets, and a minor degradation in case of the fourth, when Word2Vec embeddings are used. Finally, a comparison of the four embeddings shows that Word2Vec and dependency weight-based features outperform LSA and GloVe, in terms of their benefit to sarcasm detection. 
\end{abstract}
\section{Introduction}
\label{sec:intro}
Sarcasm is a form of verbal irony that is intended to express contempt or ridicule. Linguistic studies show that the notion of context incongruity is at the heart of sarcasm~\cite{ivanko2003context}. A popular trend in automatic sarcasm detection is semi-supervised extraction of patterns that capture the underlying context incongruity~\cite{4,22,riloff}. However, techniques to extract these patterns rely on sentiment-bearing words and may not capture nuanced forms of sarcasm. Consider the sentence `\textit{With a sense of humor like that, you could make a living as a garbage man anywhere in the country.}\footnote{All examples in this paper are actual instances from our dataset.}' The speaker makes a subtle, contemptuous remark about the sense of humor of the listener. However, absence of sentiment words makes the sarcasm in this sentence difficult to capture as features for a classifier. 

In this paper, we explore use of word embeddings to capture context incongruity in the absence of sentiment words. The intuition is that \textbf{word vector-based similarity/discordance is indicative of semantic similarity which in turn is a handle for context incongruity}. In the case of the `sense of humor' example above, the words `sense of humor' and `garbage man' are semantically dissimilar and their presence together in the sentence provides a clue to sarcasm. Hence, our set of features based on word embeddings aim to capture such semantic similarity/discordance. Since such semantic similarity is but one of the components of context incongruity \textbf{and} since existing feature sets rely on sentiment-based features to capture context incongruity, it is imperative that the two be combined for sarcasm detection. Thus, our paper deals with the question:
\begin{center}
\textit{Can word embedding-based features when augmented to features reported in prior work improve the performance of sarcasm detection?}
\end{center}

To the best of our knowledge, this is the first attempt that uses word embedding-based features to detect sarcasm. In this respect, the paper makes a simple increment to state-of-the-art but opens up a new direction in sarcasm detection research. We establish our hypothesis in case of four past works and four types of word embeddings, to show that the benefit of using word embedding-based features holds across multiple feature sets and word embeddings.
\section{Motivation}
\label{sec:motiv}
In our literature survey of sarcasm detection~\cite{sarcsurvey}, we observe that a popular trend is semi-supervised extraction of patterns with implicit sentiment. One such work is by \newcite{riloff} who give a bootstrapping algorithm that discovers a set of positive verbs and negative/undesirable situations. However, this simplification (of representing sarcasm merely as positive verbs followed by negative situation) may not capture difficult forms of context incongruity. Consider the sarcastic sentence `A woman needs a man like a fish needs bicycle'\footnote{This quote is attributed to Irina Dunn, an Australian writer (\url{https://en.wikipedia.org/wiki/Irina_Dunn}}. The sarcasm in this sentence is understood from the fact that a fish does not need bicycle - and hence, the sentence ridicules the target `a man'. However, this sentence does not contain any sentiment-bearing word. Existing sarcasm detection systems relying on sentiment incongruity (as in the case of our past work reported as \newcite{22}) may not work well in such cases of sarcasm. 

To address this, we use  semantic similarity as a handle to context incongruity. To do so, we use word vector similarity scores. Consider similarity scores (as given by Word2Vec) between two pairs of words in the sentence above:
\begin{center}
\textit{similarity(‘man’,’woman’) = 0.766}\\
\textit{similarity(‘fish’,’bicycle’) = 0.131}
\end{center}
Words in one part of this sentence (`man'  and `woman') are lot more similar than words in another part of the sentence (`fish' and `bicycle'). This semantic discordance can be a clue to presence of context incongruity. Hence, we propose features based on similarity scores between word embeddings of words in a sentence. In general, we wish to capture the most similar and most dissimilar word pairs in the sentence, and use their scores as features for sarcasm detection. 
\section{Background: Features from prior work}
\label{sec:recap}
We augment our word embedding-based features to the following four feature sets that have been reported:
\begin{enumerate}\setlength\itemsep{0cm}
\item \textbf{\newcite{liebrecht}}: They consider unigrams, bigrams and trigrams as features.
\item \textbf{\newcite{gonzalez}}: They propose two sets of features: unigrams and dictionary-based. The latter are words from a lexical resource called LIWC. We use words from LIWC that have been annotated as emotion and psychological process words, as described in the original paper. 
\item \textbf{\newcite{buschmeier}}: In addition to unigrams, they propose features such as: (a) Hyperbole (captured by three positive or negative words in a row), (b) Quotation marks and ellipsis, (c) Positive/Negative Sentiment words followed by an exclamation mark or question mark, (d) Positive/Negative Sentiment Scores followed by ellipsis (represented by a `...'), (e) Punctuation, (f) Interjections, and (g) Laughter expressions.
\item \textbf{\newcite{22}}: In addition to unigrams, they use features based on implicit and explicit incongruity. Implicit incongruity features are patterns with implicit sentiment as extracted in a pre-processing step. Explicit incongruity features consist of number of sentiment flips, length of positive and negative sub-sequences and lexical polarity.

\end{enumerate}

\section{Word Embedding-based Features}
\label{sec:newfeatures}
In this section, we now describe our word embedding-based features. We reiterate that these features will be augmented to features from prior works (described in Section~\ref{sec:recap}). 

As stated in Section ~\ref{sec:motiv}, our word embedding-based features are based on similarity scores between word embeddings. The similarity score is the cosine similarity between vectors of two words. To illustrate our features, we use our example `\textit{A woman needs a man like a fish needs a bicycle}'. The scores for all pairs of words in this sentence are given in Table~\ref{tab:example}.

\begin{table}[h!]
\centering
\begin{tabular}{p{0.8cm}p{0.7cm}p{0.8cm}p{0.8cm}p{0.7cm}p{0.8cm}}
\toprule
                 & \textbf{man} & \textbf{woman} & \textbf{fish} & \textbf{needs} & \textbf{bicycle} \\ \midrule
\textbf{man}     &   -           & \textbf{0.766}          & \textbf{0.151}         & \textbf{0.078}          & \textbf{0.229}            \\ 
\textbf{woman}   & 0.766        &     -           & 0.084         & 0.060          & 0.229            \\ 
\textbf{fish}    & 0.151       & 0.084          &       -        & 0.022          & 0.130            \\ 
\textbf{needs}   & 0.078        & 0.060          & 0.022         &       -         & 0.060            \\ 
\textbf{bicycle} & 0.229        & 0.229          & 0.130         & 0.060          &   -               \\ \bottomrule
\end{tabular}
\caption{Similarity scores for all pairs of content words in `A woman needs a man like a fish needs bicycle'}
\label{tab:example}
\end{table}

Using these similarity scores, we compute two sets of features:
\begin{enumerate}\setlength\itemsep{0cm}
\item \textbf{Unweighted similarity features (S)}: We first compute similarity scores for all pairs of words (except stop words). We then return four feature values per sentence.\footnote{These feature values consider all words in the sentence, \textit{i.e.}, the `maximum' is computed over all words}:
\begin{itemize}\setlength\itemsep{0cm}
\item Maximum score of most \textit{similar} word pair
\item Minimum score of most \textit{similar} word pair
\item Maximum score of most \textit{dissimilar} word pair
\item Minimum score of most \textit{dissimilar} word pair
\end{itemize}
For example, in case of the first feature, we consider the most similar word to every word in the sentence, and the corresponding similarity scores. These most similar word scores for each word are indicated in bold in Table~\ref{tab:example}. Thus, the first feature in case of our example would have the value 0.766 derived from the man-woman pair and the second feature would take the value 0.078 due to the needs-man pair. The other features are computed in a similar manner.
\item \textbf{Distance-weighted similarity features (WS)}: Like in the previous case, we first compute similarity scores for all pairs of words (excluding stop-words). For all similarity scores, we divide them by square of distance between the two words. Thus, the similarity between terms that are close in the sentence is weighted higher than terms which are distant from one another. Thus, for all possible word pairs, we compute four features:
\begin{itemize}\setlength\itemsep{0cm}
\item Maximum distance-weighted score of most \textit{similar} word pair
\item Minimum distance-weighted score of most \textit{similar} word pair
\item Maximum distance-weighted score of most \textit{dissimilar} word pair
\item Minimum distance-weighted score of most \textit{dissimilar} word pair
\end{itemize}
These are computed similar to unweighted similarity features.
\end{enumerate}
\section{Experiment Setup}
\label{sec:expsetup}
We create a dataset consisting of quotes on GoodReads~\footnote{\url{www.goodreads.com}}. GoodReads describes itself as `\textit{the world's largest site for readers and book recommendations.}' The website also allows users to post  quotes from books. These quotes are snippets from books labeled by the user with tags of their choice. We download quotes with the tag `sarcastic' as sarcastic quotes, and the ones with `philosophy' as non-sarcastic quotes. Our labels are based on these tags given by users. We ensure that no quote has both these tags. This results in a dataset of 3629 quotes out of which 759 are labeled as sarcastic. This skew is similar to skews observed in datasets on which sarcasm detection experiments have been reported in the past~\cite{riloff}.

We report five-fold cross-validation results on the above dataset. We use $SVM_{perf}$ by \newcite{svmperf} with $c$ as 20, $w$ as 3, and loss function as F-score optimization. This allows SVM to be learned while optimizing the F-score.

As described above, we compare features given in prior work alongside the augmented versions. This means that for each of the four papers, we experiment with four configurations:
\begin{enumerate}\setlength\itemsep{0cm}
\item Features given in paper X
\item Features given in paper X + unweighted similarity features (S) 
\item Features given in paper X + weighted similarity features (WS) 
\item Features given in paper X + S+WS (\textit{i.e.}, weighted and unweighted similarity features)
\end{enumerate}

\begin{table}[]
\centering
\begin{tabular}{p{2.5cm}p{1cm}p{1cm}p{1cm}}
\toprule
\textbf{Features} & \textbf{P} & \textbf{R} & \textbf{F} \\  \midrule
\multicolumn{4}{c}{\textbf{Baseline}}                \\ \midrule
Unigrams          & 67.2       & 78.8       & 72.53      \\ 
S   & 64.6       & 75.2       & 69.49      \\ 
WS      & 67.6       & 51.2       & 58.26      \\ 
Both              & 67         & 52.8       & 59.05      \\ \bottomrule
\end{tabular}
\caption{Performance of unigrams versus our similarity-based features using embeddings from Word2Vec}
\label{tab:res1}
\end{table}
\begin{table*}[h!]
\centering

\begin{tabular}{p{1.2cm}|lll|llll|llll|llll}
\toprule
      & \multicolumn{3}{|c|}{\textbf{LSA}}                                               &                      & \multicolumn{3}{c|}{\textbf{GloVe}}                                             &                      & \multicolumn{3}{c|}{\textbf{Dependency Weights}}                                &                      & \multicolumn{3}{c}{\textbf{Word2Vec}}                                          \\ \midrule
      & \multicolumn{1}{c}{P} & \multicolumn{1}{c}{R} & \multicolumn{1}{c}{F} & \multicolumn{1}{c}{} & \multicolumn{1}{c}{P} & \multicolumn{1}{c}{R} & \multicolumn{1}{c}{F} & \multicolumn{1}{c}{} & \multicolumn{1}{c}{P} & \multicolumn{1}{c}{R} & \multicolumn{1}{c}{F} & \multicolumn{1}{c}{} & \multicolumn{1}{c}{P} & \multicolumn{1}{c}{R} & \multicolumn{1}{c}{F} \\\midrule
\textbf{L} & 73                  & 79                  & 75.8          &                      & 73                  & 79                  & 75.8          &                      & 73                  & 79                  & 75.8          &                      & 73                    & 79                    & 75.8         \\
+S   & 81.8                 & 78.2                 & \textbf{79.95}              &                      & 81.8                 & 79.2                 & \textbf{80.47}          &                      & 81.8                 & 78.8                 & 80.27          &                      & 80.4                  & 80                    & \textbf{80.2}         \\
+WS    & 76.2                 & 79.8                 & 77.9          &                      & 76.2                 & 79.6                 & 77.86          &                      & 81.4                 & 80.8                 & 81.09          &                      & 80.8                  & 78.6                  & 79.68         \\
+S+WS & 77.6                 & 79.8                 & 78.68          &                      & 74                  & 79.4                 & 76.60          &                      & 82                  & 80.4                 & \textbf{81.19}          &                      & 81.6                  & 78.2                  & 79.86         \\
 \midrule
\textbf{G} & 84.8                 & 73.8                 & 78.91           &                      & 84.8                 & 73.8                 & 78.91           &                      & 84.8                 & 73.8                 & \textbf{78.91}           &                      & 84.8                  & 73.8                  & \textbf{78.91}         \\
+S   & 84.2                 & 74.4                 & \textbf{79}          &                      & 84                  & 72.6                 & 77.8          &                      & 84.4                 & 72                  & 77.7          &                      & 84                    & 72.8                  & 78                    \\
+WS    & 84.4                 & 73.6                 & 78.63          &                      & 84                  & 75.2                 & \textbf{79.35}          &                      & 84.4                 & 72.6                 & 78.05          &                      & 83.8                  & 70.2                  & 76.4         \\
+S+WS & 84.2                 & 73.6                 & 78.54          &                      & 84                  & 74                  & 78.68           &                      & 84.2                 & 72.2                 & 77.73          &                      & 84                    & 72.8                  & 78                    \\\midrule
\textbf{B} & 81.6                 & 72.2                 & 76.61          &                      & 81.6                 & 72.2                 & 76.61          &                      & 81.6                 & 72.2                 & 76.61          &                      & 81.6                 & 72.2                 & 76.61         \\
+S   & 78.2                 & 75.6                 & \textbf{76.87}          &                      & 80.4                 & 76.2                 &\textbf{78.24}          &                      & 81.2                 & 74.6                 & \textbf{77.76}          &                      & 81.4                  & 72.6                  & 76.74         \\
+WS    & 75.8                 & 77.2                 & 76.49          &                      & 76.6                 & 77                  & 76.79          &                      & 76.2                 & 76.4                 & 76.29          &                      & 81.6                  & 73.4                  & 77.28         \\
+S+WS & 74.8                 & 77.4                 & 76.07          &                      & 76.2                 & 78.2                 & 77.18          &                      & 75.6                 & 78.8                 & 77.16          &                      & 81                    & 75.4                  & \textbf{78.09}         \\\midrule
\textbf{J} & 85.2                  & 74.4                  & 79.43           &                      & 85.2                  & 74.4                  & 79.43            &                      & 85.2                  & 74.4                  & 79.43           &                      & 85.2                  & 74.4                  & 79.43         \\
+S   & 84.8                 & 73.8                 & 78.91           &                      & 85.6                 & 74.8                 & 79.83          &                      & 85.4                 & 74.4                 & 79.52          &                      & 85.4                  & 74.6                  & \textbf{79.63}               \\
+WS    & 85.6                 & 75.2                 & \textbf{80.06}          &                      & 85.4                 & 72.6                 & 78.48          &                      & 85.4                 & 73.4                 & 78.94           &                      & 85.6                  & 73.4                  & 79.03         \\
+S+WS & 84.8                 & 73.6                 & 78.8          &                      & 85.8                 & 75.4                 & \textbf{80.26}          &                      & 85.6                 & 74.4                 & \textbf{79.6}               &                      & 85.2                  & 73.2                  & 78.74        \\ \bottomrule
\end{tabular}
\caption{Performance obtained on augmenting word embedding features to features from four prior works, for four word embeddings; L: Liebrecht et al. (2013), G: Gonz\'alez-Ib\'anez et al. (2011a), B: Buschmeier et al. (2014) , J: Joshi et al. (2015)}
\label{tab:res2}
\end{table*}
We experiment with four types of word embeddings:
\begin{enumerate}
\item \textbf{LSA}: This approach was reported in \newcite{Landauer97asolution}. We use pre-trained word embeddings based on LSA\footnote{\url{ http://www.lingexp.uni-tuebingen.de/z2/LSAspaces/}}. The vocabulary size is 100,000. 
\item \textbf{GloVe}: We use pre-trained vectors avaiable from the GloVe project\footnote{\url{http://nlp.stanford.edu/projects/glove/}}. The vocabulary size in this case is 2,195,904.  
\item \textbf{Dependency Weights}: We use pre-trained vectors\footnote{~\url{https://levyomer.wordpress.com/2014/04/25/dependency-based-word-embeddings/}} weighted using dependency distance, as given in ~\newcite{depwords}. The vocabulary size is 174,015. 
\item \textbf{Word2Vec}: use pre-trained Google word vectors. These were trained using Word2Vec tool \footnote{\url{https://code.google.com/archive/p/Word2Vec/}} on the Google News corpus. The vocabulary size for Word2Vec is 3,000,000. To interact with these pre-trained vectors, as well as compute various features, we use gensim library \cite{rehurek_lrec}.
\end{enumerate}
To interact with the first three pre-trained vectors, we use scikit library~\cite{scikit}.


\section{Results}
\label{sec:results}
Table~\ref{tab:res1} shows performance of sarcasm detection when our word embedding-based features are used on their own \textit{i.e}, not as augmented features. The embedding in this case is Word2Vec. The four rows show baseline sets of features: unigrams, unweighted similarity using word embeddings (S), weighted similarity using word embeddings (WS) and both (\textit{i.e.}, unweighted plus weighted similarities using word embeddings). Using only unigrams as features gives a F-score of 72.53\%, while only unweighted and weighted features gives F-score of 69.49\% and 58.26\% respectively. This \textbf{validates our intuition that word embedding-based features alone are not sufficient, and should be augmented with other features}. 

Following this, we show performance using features presented in four prior works: \newcite{buschmeier}, \newcite{liebrecht}, \newcite{22} and \newcite{gonzalez}, and compare them with augmented versions in Table~\ref{tab:res2}. 

Table~\ref{tab:res2} shows results for four kinds of word embeddings. All entries in the tables are \textbf{higher than the simple unigrams baseline}, \textit{i.e.}, F-score for each of the four is higher than unigrams - highlighting that these are better features for sarcasm detection than simple unigrams. Values in bold indicate the best F-score for a given prior work-embedding type combination. In case of \newcite{liebrecht} for Word2Vec, the overall improvement in F-score is 4\%. Precision increases by 8\% while recall remains nearly unchanged. For features given in \newcite{gonzalez}, there is a negligible degradation of 0.91\% when word embedding-based features based on Word2Vec are used. For \newcite{buschmeier} for Word2Vec, we observe an improvement in F-score from 76.61\% to 78.09\%. Precision remains nearly unchanged while recall increases. In case of \newcite{22} and Word2Vec, we observe a slight improvement of 0.20\% when unweighted (S) features are used. This shows that word embedding-based features are useful, across four past works for Word2Vec.

Table~\ref{tab:res2} also shows that the improvement holds across the four word embedding types as well. The maximum improvement is observed in case of \newcite{liebrecht}. It is around 4\% in case of LSA, 5\% in case of GloVe, 6\% in case of Dependency weight-based and 4\% in case of Word2Vec. These improvements are not directly comparable because the four embeddings have different vocabularies (since they are trained on different datasets) and vocabulary sizes, their results cannot be directly compared. 

Therefore, we take an intersection of the vocabulary (\textit{i.e.}, the subset of words present in all four embeddings) and repeat all our experiments using these intersection files. The vocabulary size of these intersection files (for all four embeddings) is 60,252. Table ~\ref{tab:res3} shows the average increase in F-score when a given word embedding and a word embedding-based feature is used, with the intersection file as described above. These gain values are lower than in the previous case. This is because these are the values in case of the intersection versions - which are subsets of the complete embeddings. Each gain value is averaged over the four prior works. Thus, when unweighted similarity (+S) based features computed using LSA are augmented to features from prior work, an average increment of 0.835\% is obtained over the four prior works.  The values allow us to compare the benefit of using these four kinds of embeddings. In case of unweighted similarity-based features, dependency-based weights give the maximum gain (0.978\%). In case of weighted similarity-based features and `+S+WS', Word2Vec gives the maximum gain (1.411\%). Table~\ref{tab:res4} averages these values over the three types of word embedding-based features. Using Dependency-based and Word2Vec embeddings results in a higher improvement in F-score (1.048\% and 1.143\% respectively) as compared to others.

\begin{table}[]
\centering
\begin{tabular}{p{1cm}p{1.3cm}p{1.1cm}p{1.1cm}p{1.3cm}}
\toprule
& \textbf{Word2Vec} & \textbf{LSA} & \textbf{GloVe} & \textbf{Dep. Wt.} \\ \midrule
+S        & 0.835             & 0.86         & 0.918          & \textbf{0.978}        \\
+WS       & \textbf{1.411}             & 0.255        & 0.192          & 1.372        \\
+S+WS     & \textbf{1.182}             & 0.24         & 0.845          & 0.795       \\ \bottomrule
\end{tabular}
\caption{Average gain in F-Scores obtained by using intersection of the four word embeddings, for three word embedding feature-types, augmented to four prior works; Dep. Wt. indicates vectors learned from dependency-based weights}
\label{tab:res3}
\end{table}
\begin{table}[]
\centering
\begin{tabular}{cc}
\toprule
\textbf{Word Embedding} & \textbf{Average F-score Gain} \\ \midrule
LSA            & 0.452                   \\
Glove          & 0.651                   \\
Dependency     & 1.048                   \\
Word2Vec       & 1.143   \\\bottomrule               
\end{tabular}
\caption{Average gain in F-scores for the four types of word embeddings; These values are computed for a subset of these embeddings consisting of words common to all four}
\label{tab:res4}
\end{table}
\section{Error Analysis}
\label{sec:erroranal}
Some categories of errors made by our system are:
\begin{enumerate}\setlength\itemsep{0cm}
\item \textbf{Embedding issues due to incorrect senses}: Because words may have multiple senses, some embeddings lead to error, as in `\textit{Great. Relationship advice from one of America's most wanted.}'. 
\item \textbf{Contextual sarcasm}: Consider the sarcastic quote `\textit{Oh, and I suppose the apple ate the cheese}'. The similarity score between `apple' and `cheese' is 0.4119. This comes up as the maximum similar pair. The most dissimilar pair is `suppose' and `apple' with similarity score of 0.1414. The sarcasm in this sentence can be understood only in context of the complete conversation that it is a part of.
\item \textbf{Metaphors in non-sarcastic text}: Figurative language may compare concepts that are not directly related but still have low similarity. Consider the non-sarcastic quote `\textit{Oh my love, I like to vanish in you like a ripple vanishes in an ocean - slowly, silently and endlessly}'. Our system incorrectly predicts this as sarcastic.
\end{enumerate}
\section{Related Work}
\label{sec:relwork}
Early sarcasm detection research focused on speech~\cite{1} and lexical features~\cite{2}. Several other features have been proposed ~\cite{2,22,23,liebrecht,gonzalez,15,26,19,5,6,7}. Of particular relevance to our work are papers that aim to first extract patterns relevant to sarcasm detection. \newcite{4} use a semi-supervised approach that extracts sentiment-bearing patterns for sarcasm detection. \newcite{22} extract phrases corresponding to implicit incongruity i.e. the situation where sentiment is expressed without use of sentiment words. \newcite{riloff} describe a bootstrapping algorithm that iteratively discovers a set of positive verbs and negative situation phrases, which are later used in a sarcasm detection algorithm. \newcite{3} also perform semi-supervised extraction of patterns for sarcasm detection. The only prior work which uses word embeddings for a related task of sarcasm detection is by \newcite{ghoshsarcastic}. They model sarcasm detection as a word sense disambiguation task, and use embeddings to identify whether a word is used in the sarcastic or non-sarcastic sense. Two sense vectors for every word are created: one for literal sense and one for sarcastic sense. The final sense is determined based on the similarity of these sense vectors with the sentence vector.


\section{Conclusion}
\label{sec:concl}
This paper shows the benefit of features based on word embedding for sarcasm detection. We experiment with four past works in sarcasm detection, where we augment our word embedding-based features to their sets of features. Our features use the similarity score values returned by word embeddings, and are of two categories: similarity-based (where we consider maximum/minimum similarity score of most similar/dissimilar word pair respectively), and weighted similarity-based (where we weight the maximum/minimum similarity scores of most similar/dissimilar word pairs with the linear distance between the two words in the sentence). We experiment with four kinds of word embeddings: LSA, GloVe, Dependency-based and Word2Vec. In case of Word2Vec, for three of these past feature sets to which our features were augmented, we observe an improvement in F-score of at most 5\%. Similar improvements are observed in case of other word embeddings. A comparison of the four embeddings shows that Word2Vec and dependency weight-based features outperform LSA and GloVe.

This work opens up avenues for use of word embeddings for sarcasm classification. Our word embedding-based features may work better if the similarity scores are computed for a subset of words in the sentence, or using weighting based on syntactic distance instead of linear distance as in the case of our weighted similarity-based features.

\bibliographystyle{emnlp2016}
\bibliography{acl2016}

\begin{thebibliography}{}

\bibitem[\protect\citename{Buschmeier \bgroup et al.\egroup }2014]{buschmeier}
Konstantin Buschmeier, Philipp Cimiano, and Roman Klinger.
\newblock 2014.
\newblock An impact analysis of features in a classification approach to irony
  detection in product reviews.
\newblock In {\em Proceedings of the 5th Workshop on Computational Approaches
  to Subjectivity, Sentiment and Social Media Analysis}, pages 42--49.

\bibitem[\protect\citename{Davidov \bgroup et al.\egroup }2010]{4}
Dmitry Davidov, Oren Tsur, and Ari Rappoport.
\newblock 2010.
\newblock Semi-supervised recognition of sarcastic sentences in twitter and
  amazon.
\newblock In {\em Proceedings of the Fourteenth Conference on Computational
  Natural Language Learning}, pages 107--116. Association for Computational
  Linguistics.

\bibitem[\protect\citename{Ghosh \bgroup et al.\egroup }2015]{ghoshsarcastic}
Debanjan Ghosh, Weiwei Guo, and Smaranda Muresan.
\newblock 2015.
\newblock Sarcastic or not: Word embeddings to predict the literal or sarcastic
  meaning of words.
\newblock In {\em EMNLP}.

\bibitem[\protect\citename{Gonz{\'a}lez-Ib{\'a}nez \bgroup et al.\egroup
  }2011a]{gonzalez}
Roberto Gonz{\'a}lez-Ib{\'a}nez, Smaranda Muresan, and Nina Wacholder.
\newblock 2011a.
\newblock Identifying sarcasm in twitter: a closer look.
\newblock In {\em Proceedings of the 49th Annual Meeting of the Association for
  Computational Linguistics: Human Language Technologies: short papers-Volume
  2}, pages 581--586. Association for Computational Linguistics.

\bibitem[\protect\citename{Gonz{\'a}lez-Ib{\'a}nez \bgroup et al.\egroup
  }2011b]{6}
Roberto Gonz{\'a}lez-Ib{\'a}nez, Smaranda Muresan, and Nina Wacholder.
\newblock 2011b.
\newblock Identifying sarcasm in twitter: a closer look.
\newblock In {\em Proceedings of the 49th Annual Meeting of the Association for
  Computational Linguistics: Human Language Technologies: short papers-Volume
  2}, pages 581--586. Association for Computational Linguistics.

\bibitem[\protect\citename{Ivanko and Pexman}2003]{ivanko2003context}
Stacey~L Ivanko and Penny~M Pexman.
\newblock 2003.
\newblock Context incongruity and irony processing.
\newblock {\em Discourse Processes}, 35(3):241--279.

\bibitem[\protect\citename{Joachims}2006]{svmperf}
Thorsten Joachims.
\newblock 2006.
\newblock Training linear svms in linear time.
\newblock In {\em Proceedings of the 12th ACM SIGKDD international conference
  on Knowledge discovery and data mining}, pages 217--226. ACM.

\bibitem[\protect\citename{Joshi \bgroup et al.\egroup }2015]{22}
Aditya Joshi, Vinita Sharma, and Pushpak Bhattacharyya.
\newblock 2015.
\newblock Harnessing context incongruity for sarcasm detection.
\newblock In {\em Proceedings of the 53rd Annual Meeting of the Association for
  Computational Linguistics and the 7th International Joint Conference on
  Natural Language Processing}, volume~2, pages 757--762.

\bibitem[\protect\citename{Joshi \bgroup et al.\egroup }2016]{sarcsurvey}
Aditya Joshi, Pushpak Bhattacharyya, and Mark~James Carman.
\newblock 2016.
\newblock Automatic sarcasm detection: A survey.
\newblock {\em arXiv preprint arXiv:1602.03426}.

\bibitem[\protect\citename{Khattri \bgroup et al.\egroup }2015]{23}
Anupam Khattri, Aditya Joshi, Pushpak Bhattacharyya, and Mark~James Carman.
\newblock 2015.
\newblock Your sentiment precedes you: Using an author’s historical tweets to
  predict sarcasm.
\newblock In {\em 6th Workshop on Computational Approaches to Subjectivity,
  Sentiment and Social Media Analysis (WASSA)}, page~25.

\bibitem[\protect\citename{Kreuz and Caucci}2007]{2}
Roger~J Kreuz and Gina~M Caucci.
\newblock 2007.
\newblock Lexical influences on the perception of sarcasm.
\newblock In {\em Proceedings of the Workshop on computational approaches to
  Figurative Language}, pages 1--4. Association for Computational Linguistics.

\bibitem[\protect\citename{Landauer and Dumais}1997]{Landauer97asolution}
Thomas~K Landauer and Susan~T. Dumais.
\newblock 1997.
\newblock A solution to plato’s problem: The latent semantic analysis theory
  of acquisition, induction, and representation of knowledge.
\newblock {\em PSYCHOLOGICAL REVIEW}, 104(2):211--240.

\bibitem[\protect\citename{Levy and Goldberg}2014]{depwords}
Omer Levy and Yoav Goldberg.
\newblock 2014.
\newblock Dependency-based word embeddings.
\newblock In {\em Proceedings of the 52nd Annual Meeting of the Association for
  Computational Linguistics, {ACL} 2014, June 22-27, 2014, Baltimore, MD, USA,
  Volume 2: Short Papers}, pages 302--308.

\bibitem[\protect\citename{Liebrecht \bgroup et al.\egroup }2013]{liebrecht}
CC~Liebrecht, FA~Kunneman, and APJ van~den Bosch.
\newblock 2013.
\newblock The perfect solution for detecting sarcasm in tweets\# not.

\bibitem[\protect\citename{Pedregosa \bgroup et al.\egroup }2011]{scikit}
Fabian Pedregosa, Ga{\"e}l Varoquaux, Alexandre Gramfort, Vincent Michel,
  Bertrand Thirion, Olivier Grisel, Mathieu Blondel, Peter Prettenhofer, Ron
  Weiss, Vincent Dubourg, et~al.
\newblock 2011.
\newblock Scikit-learn: Machine learning in python.
\newblock {\em The Journal of Machine Learning Research}, 12:2825--2830.

\bibitem[\protect\citename{Rakov and Rosenberg}2013]{15}
Rachel Rakov and Andrew Rosenberg.
\newblock 2013.
\newblock " sure, i did the right thing": a system for sarcasm detection in
  speech.
\newblock In {\em INTERSPEECH}, pages 842--846.

\bibitem[\protect\citename{{\v R}eh{\r u}{\v r}ek and Sojka}2010]{rehurek_lrec}
Radim {\v R}eh{\r u}{\v r}ek and Petr Sojka.
\newblock 2010.
\newblock {Software Framework for Topic Modelling with Large Corpora}.
\newblock In {\em {Proceedings of the LREC 2010 Workshop on New Challenges for
  NLP Frameworks}}, pages 45--50, Valletta, Malta, May. ELRA.
\newblock \url{http://is.muni.cz/publication/884893/en}.

\bibitem[\protect\citename{Reyes \bgroup et al.\egroup }2012]{7}
Antonio Reyes, Paolo Rosso, and Davide Buscaldi.
\newblock 2012.
\newblock From humor recognition to irony detection: The figurative language of
  social media.
\newblock {\em Data \& Knowledge Engineering}, 74:1--12.

\bibitem[\protect\citename{Riloff \bgroup et al.\egroup }2013]{riloff}
Ellen Riloff, Ashequl Qadir, Prafulla Surve, Lalindra De~Silva, Nathan Gilbert,
  and Ruihong Huang.
\newblock 2013.
\newblock Sarcasm as contrast between a positive sentiment and negative
  situation.
\newblock In {\em EMNLP}, pages 704--714.

\bibitem[\protect\citename{Tepperman \bgroup et al.\egroup }2006]{1}
Joseph Tepperman, David~R Traum, and Shrikanth Narayanan.
\newblock 2006.
\newblock " yeah right": sarcasm recognition for spoken dialogue systems.
\newblock In {\em INTERSPEECH}. Citeseer.

\bibitem[\protect\citename{Tsur \bgroup et al.\egroup }2010]{3}
Oren Tsur, Dmitry Davidov, and Ari Rappoport.
\newblock 2010.
\newblock Icwsm-a great catchy name: Semi-supervised recognition of sarcastic
  sentences in online product reviews.
\newblock In {\em ICWSM}.

\bibitem[\protect\citename{Veale and Hao}2010]{5}
Tony Veale and Yanfen Hao.
\newblock 2010.
\newblock Detecting ironic intent in creative comparisons.
\newblock In {\em ECAI}, volume 215, pages 765--770.

\bibitem[\protect\citename{Wallace \bgroup et al.\egroup }2014]{19}
Byron~C Wallace, Laura~Kertz Do~Kook~Choe, and Eugene Charniak.
\newblock 2014.
\newblock Humans require context to infer ironic intent (so computers probably
  do, too).
\newblock In {\em Proceedings of the Annual Meeting of the Association for
  Computational Linguistics (ACL)}, pages 512--516.

\bibitem[\protect\citename{Wallace}2015]{26}
Byron~C Wallace.
\newblock 2015.
\newblock Sparse, contextually informed models for irony detection: Exploiting
  user communities,entities and sentiment.
\newblock In {\em ACL}.

\end{thebibliography}
\end{document}